\documentclass{article}
\usepackage{spconf,amsmath,graphicx}
\usepackage{amsthm,amssymb}
\usepackage{mathrsfs}

\usepackage{enumitem}
\setlist{nosep, leftmargin=14pt}
\usepackage{mwe}
\usepackage{multicol}
\usepackage{multirow}
\usepackage{bbding}
\usepackage{arydshln}
\usepackage{subcaption}
\usepackage{booktabs}
\usepackage{amsmath}
\usepackage{xcolor}
\definecolor{cvprblue}{rgb}{0.21,0.49,0.74}
\usepackage[colorlinks, urlcolor=black, citecolor=cvprblue]{hyperref}

\title{Annotation-Efficient Polyp Segmentation via Active Learning}
\name{\begin{tabular}{c}Duojun Huang$^1$, Xinyu Xiong$^1$, De-Jun Fan$^1$, Feng Gao$^2$, Xiao-Jian Wu$^2$, Guanbin Li$^{1, 3*}$\thanks{$^*$ Corresponding Author.}\end{tabular}}
\address{$^1$School of Computer Science and Engineering, Sun Yat-sen University, Guangzhou, China\\
$^2$Department of General Surgery (Department of Colorectal Surgery), \\The Sixth Affiliated Hospital, Sun Yat-sen University, Guangzhou, China\\
$^3$Research Institute, Sun Yat-sen University, Shenzhen, China
}

\begin{document}
%
\maketitle

\begin{abstract}
Deep learning-based techniques have proven effective in polyp segmentation tasks when provided with sufficient pixel-wise labeled data. However, the high cost of manual annotation has created a bottleneck for model generalization. To minimize annotation costs, we propose a deep active learning framework for annotation-efficient polyp segmentation. In practice, we measure the uncertainty of each sample by examining the similarity between features masked by the prediction map of the polyp and the background area. Since the segmentation model tends to perform weak in samples with indistinguishable features of foreground and background areas, uncertainty sampling facilitates the fitting of under-learning data. Furthermore, clustering image-level features weighted by uncertainty identify samples that are both uncertain and representative. To enhance the selectivity of the active selection strategy, we propose a novel unsupervised feature discrepancy learning mechanism. The selection strategy and feature optimization work in tandem to achieve optimal performance with a limited annotation budget. Extensive experimental results have demonstrated that our proposed method achieved state-of-the-art performance compared to other competitors on both a public dataset and a large-scale in-house dataset.

\keywords{Active Learning, Polyp Segmentation, Uncertainty Estimation}
\end{abstract}

\section{Introduction}

Colorectal cancer (CRC) is a significant cause of morbidity and mortality, ranking third among all cancers and posing a serious threat to human health~\cite{siegel2020crcstat,zhang2022lesion}. Polyp segmentation plays a crucial role in detecting and treating CRC in its early stages, as it is a significant precursor of cancer. Recent deep learning-based techniques~\cite{MICCAI20_ACSNet,wang2024unveiling,xiong2023unpaired,10097456,resnet,zhang2024masked} have demonstrated their effectiveness in image representation and segmentation tasks. However, the success of these methods relies on the availability of sufficient well-annotated data, which is labor-intensive to collect and annotate for polyp segmentation tasks for several reasons: (1) The annotation requires experts with professional medical knowledge to locate the position of polyps; (2) Colonoscopy image datasets are generally collected from video sequences~\cite{li2023hybridvps,MICCAI22_SemiVPS,zhang2024distribution}, requiring human experts to examine each frame to identify polyps and precisely annotate each pixel in the polyp area; (3) Polyps vary in color, shape, and location in the colon, which needs to incorporate image data from diverse scenarios to guarantee the model's generalization ability.

Recently, active learning (AL) methods~\cite{yang2017suggestive,wang2021annotation,li2020attention,huang2023divide} have been proposed to reduce the burden of annotation and suggest annotation on the most informative data in several medical tasks. For instance,~\cite{yang2017suggestive} proposes a novel fully convolutional neural network to achieve similarity between images in the labeled and unlabeled set and suggest annotation on valuable samples. However, training multiple networks required for the selection strategy results in increased computation costs. AECC~\cite{wang2021annotation} aims to select informative areas by utilizing extracted feature maps of deep models for the cell counting task. However, computing the similarity between unlabeled and labeled samples is not efficient and introduces additional tuning of the related hyperparameters. Most of the previous active learning methods ignore the information contained in unlabeled data, which could be distilled to enhance the feature representation of the model.

\begin{figure*}[tp]
    \centering
    \includegraphics[width=12cm]{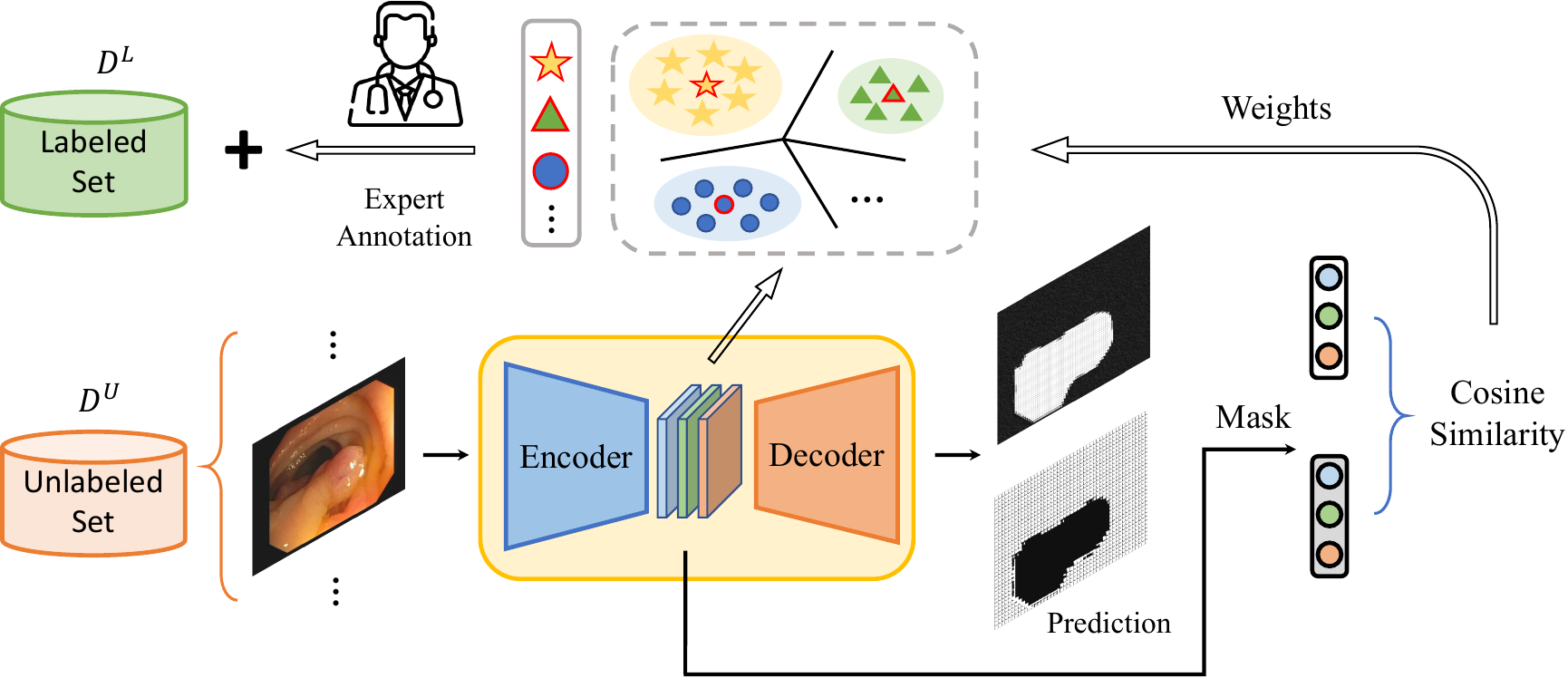}
    \caption{The work flow of our proposed active learning framework. Given an unlabeled sample $x^u$ from the unlabeled set $D^U$, our goal is to accurately measure the value of $x^u$, and high value one will be selected for expert annotation.}
    \label{fig:network}
\end{figure*}

The above observation motivates us to propose an active learning framework for polyp segmentation tasks, aimed at minimizing annotation costs by distilling valuable information from both labeled and unlabeled samples. Compared to existing methods that relies on diversity sampling or uncertainty sampling seperately, our approach selects representative and uncertain samples in a unified framework. Since the segmentation model tends to perform weak in samples with indistinguishable features of foreground and background areas, we first measure uncertainty by the similarity between features masked by the prediction map of the polyp and background area. To obtain image-level features, we use the output of the encoder in the segmentation model. Further, we perform feature clustering weighted by uncertainty to identify samples that are both uncertain and representative. Our contributions are summarized as follows:
\begin{itemize}[noitemsep, nolistsep]

\item A simple but effective active sampling query strategy that jointly captures the uncertainty and diversity of samples for the medical segmentation task. 
\item A novel unsupervised feature discrepancy learning approach that enhances the distinctiveness of features from different classes extracted by the segmentation network, which further facilitates the sample selection process. 
\item Better performance compared with other active learning methods.
\end{itemize}

\section{Method}
The overview of our active learning framework is depicted in Fig.~\ref{fig:network}.   
In each active learning loop, $B$ valueable samples from unlabeled dataset $D_{U}$ will be retrieve by our framework. These selected samples will be annotated by human experts and moved into labeled dataset $D_{L}$, followed by the model fine-tuning with the updated datasets. The procedure is repeated until the annotation budget is exhausted.

In the next, we will first present how to select valuable samples through uncertain-weighted clustering (Section~\ref{sec:query}). Subsequently, we will delve into utilizing unlabeled samples to enhance feature representations of the segmentation model for better selection (Section~\ref{sec:unsup}).

\begin{figure*}[tp]
    \centering
    \includegraphics[width=12cm]{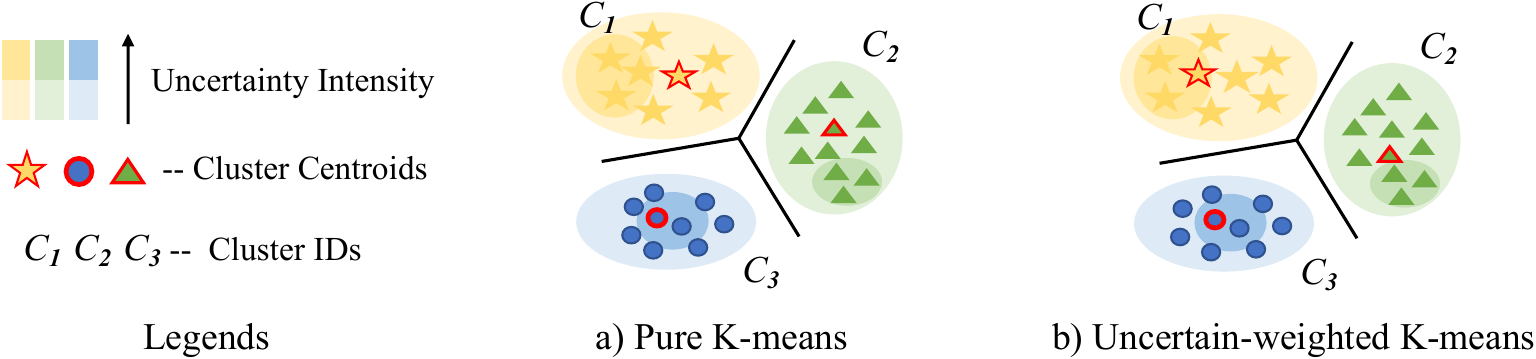}
    \caption{Comparison between the a) traditional pure K-means and b) our proposed uncertain-weighted clustering algorithm.}
    \label{fig:kmeans}
\end{figure*}

\subsection{Sampling Query Strategy}
\label{sec:query}

Under the semantic segmentation task, the extracted features of different classes (e.g. foreground lesion v.s. background tissue) for well-learned samples should be easy to discriminate. The segmentation model tends to produce a significant prediction bias for samples with similar features across the polyp and background classes~\cite{TPAMI21_COD}. Intuitively, we measure the uncertainty of the sample by the similarity of features masked by the prediction map of the task model. Specifically, we first calculate the polyp and background class feature map of each unlabeled image $x^{u}$: 
\begin{equation}
    F_{c}(x^u) = \mathop{Avg} (\hat{y}_{c}^u  \otimes f_{E}(x^{u})), 
\end{equation}
where $f_{E}(\cdot)$ and $\hat{y}_{c}^u$ denote the encoder network of the segmentation model and the prediction probability map respectively. $\otimes$ denotes element-wise multiplication for masking the polyp or the background area of the feature map. $\mathop{Avg}$ indicates the average pooling operation to transform the feature map into a vector. The class $c\in\{0, 1\}$ and 1 and 0 denote the polyp and the background class respectively. Noted that $\hat{y}_{c}^u$ is downsampled to the same size as $f_{E}(x^{u})$. Therefore, the uncertainty score of each unlabeled instance $x^{u}$ can be calculated as follows: 
\begin{equation}
    I(x^{u}) = \mathop{cos}(F_{1}(x^u), F_{0}(x^u)), 
\end{equation}
where $\mathop{cos}$ denotes the cosine similarity. As indicated by the recent state-of-the-arts~\cite{ash2019BADGE,parvaneh2022alpha}, an optimal active learning strategy should consider both the uncertainty and diversity during the selection process. To explore the representative samples, we start from the observation that a quantity of samples shares similar image-level characteristics such as surrounding, lightning, etc. 

Therefore, we propose to achieve diverse sampling based on the clustering of the image-level features extracted by the deep neural networks. Thus we can actively query the centroid of each cluster to reduce the redundancy of sampled data subset. Then, the image-level-feature is obtained by the output of the encoder network of the task model for each unlabeled sample $x^{u}$, which can be expressed as follows: 
\begin{equation}
    F(x) = \mathop{Avg}(f_{E}(x)), 
\end{equation}
where $f_{E}(\cdot)$ denotes the encoder network of common segmentation model. To jointly capture the uncertainty and representativeness of sample, we combine these two metrics by the image-level feature clustering weighted by the uncertainty score. The clustering algorithm is required to divide all the unlabeled data $\{x^{u}\}$ into $K$ groups of instances, creating a partition $S=\{S_{1}, S_{2}, ..., S_{K} \}$. With the informativeness score as the feature embedding weight, the optimization objective of the algorithm can be expressed as follows: 
\begin{equation}\label{kmeans}
\mathop{argmin}\limits_{\mathcal{S}} \sum_{k}  \frac{1}{\left\vert S_{k} \right\vert} \sum_{x\in S_{k}} I(x) \cdot \left\vert \left\vert F(x)-\mu_{k} \right\vert\right\vert^{2}, 
\end{equation}
where $\mu_{k}$ denotes the cluster centroid of the $k$-th data subset. It can be indicated from Eq.~\ref{kmeans} that the samples with higher uncertainty score will have a greater effect on the optimization of the clustering algorithm, making the density of uncertain samples increases around the cluster centroids. In each sampling cycle, $K$ is set to be the same as the given annotation budget $B$. After the clustering algorithm is optimized till coverage, we select the sample which is the nearest to the cluster centroid from each cluster. With the clustering weighted by the sample uncertainty, the uncertainty and representativeness can be jointly captured to ensure effective suggestive annotation on the most informative samples for the model training.

\subsection{Feature Discrepancy Learning}
\label{sec:unsup}

In Section~\ref{sec:query}, we outlined the methodology for selecting valuable samples from the unlabeled set to boost the segmentation performance. Nevertheless, the untapped potential of the remaining unlabeled samples remains to be fully explored. Motivated by the success of unsupervised representation learning~\cite{ICLR18_rl}, we introduce a novel feature discrepancy learning strategy that aligns seamlessly with our sampling query strategy, aiming to improve performance further within a limited annotation budget.

Specifically, during the training stage, we magnify the differentiation between the features of polyp and background for unlabeled samples by the feature discrepancy loss $\mathcal{L}_{cts}$: 
\begin{equation}\label{loss_u}
\mathcal{L}_{fdl} = \mathbb{E}_{(x,) \sim D_{U}}  max[0, \mathop{cos}(F_{1}(x^u), F_{0}(x^u))-\delta], 
\end{equation}

where $\delta$ denotes the margin to modulate the optimization intensity. We set $\delta$ to 0.2 in all experiments. Through accentuating the differentiation between the polyp and background classes, our selection strategy effectively discerns samples exhibiting higher inter-class similarity. Typically, these samples show subpar segmentation performance~\cite{fan2020pranet}, rendering them highly valuable in our selection process.

\subsection{Loss Function}
To train the segmentation network, we combine the binary cross entropy loss $\mathcal{L}_{BCE}$ and the dice loss $\mathcal{L}_{Dice}$ to supervise labeled samples: 
\begin{equation}\label{suploss}
\mathcal{L}_{seg} = \mathbb{E}_{(x,y) \sim D_{L}} \big [ \mathcal{L}_{BCE}(\hat{y}, y) + \mathcal{L}_{Dice}(\hat{y}, y) \big ],
\end{equation}
where $\hat{y}$ is the prediction of the network w.r.t sample $x$. 

Our overall training loss function consists of the supervised segmentation loss and the unsupervised feature discrepancy loss $\mathcal{L}_{fdl}$:
\begin{equation}\label{loss_all}
\mathcal{L} = \mathcal{L}_{seg} + \lambda_{c} \mathcal{L}_{fdl} , 
\end{equation}
where $\lambda_{c}$ is the trade off weight. We set $\lambda_{c}$ to 0.1 in all experiments.

\section{Experiments}

\subsection{Experimental Setup}

\textbf{Datasets and Evaluation Metrics.} To evaluate the performance of our proposed active learning framework, we choose two polyp segmentation datasets collected from different institutions. The CVC-ClinicDB \cite{cvc-clinicdb} and in-house dataset contain 612 and 1,100 images respectively. We follow previous works \cite{fan2020pranet} to divide CVC-ClinicDB into the training set and test set with 550 and 62 images respectively, and split the our in-house dataset into 1000 images for training and 100 images for testing. Each image is resized to 352 $\times$ 352 as input of the network. We use ``mIoU'' and ``Dice'' metrics to evaluate the performance of the segmentation model. 

\textbf{Implementation Details. } For active learning, we set the number of sampling rounds $R$ as 5 and the annotation budget of each sampling round $B$ as 30 across different datasets. We choose U-Net~\cite{ronneberger2015unet} as the baseline segmentation model. The batch size is set to 16 and Adam is used as the optimizer with a learning rate of $1 \times 10^{-4}$. After the sampling step in each active learning loop, the model is trained for 5 and 3 epochs for CVC-ClinicDB~\cite{cvc-clinicdb} and in-house dataset respectively. 

\textbf{Baseline Methods.} We compare the proposed method with the baseline selection methods involving random selection (Random), entropy sampling (Entropy), Coreset~\cite{sener2018coreset}, and ASA~\cite{li2020attention}.

\begin{figure}[tp]
    \centering
    \includegraphics[width=0.4\textwidth]{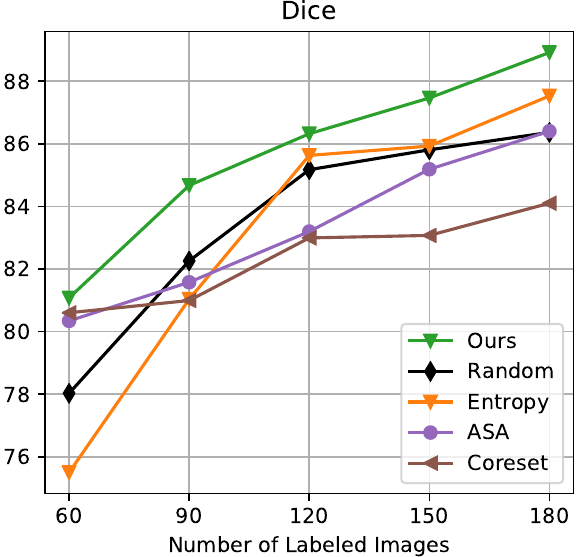}
    \caption{ Performance comparison between different method on CVC-ClinicDB dataset. For simplicity, the full performance plots for the anonymous dataset are provided in the supplemental material. }
    \label{fig:cvc}
\end{figure}

\subsection{Experimental Results}
\textbf{Result Analysis.} As depicted in Table \ref{table1} and Fig. \ref{fig:cvc}, our proposed method consistently outperforms the baseline and state-of-the-art active learning methods in both metrics with different annotation budgets. In terms of mIoU and Dice, we achieve 1.1\% and 1.4\% performance gain over the nearest competitor Entropy in the CVC-ClinicDB dataset in the final round. For the same scenario, we outperform the random baseline by 2.1\% and 2.5\% respectively. For the anonymous dataset, we achieve 1.2\% and 1.8\% performance gains over the recently proposed SOTA method ASA specifically designed for commonly-seen medical segmentation tasks. The comparison indicates the informativeness of the samples selected by our method, which can be trained to enhance the representation of the task model more sufficiently robustly. And it also verified the superiority of jointly capturing the uncertainty and diversity of samples for the segmentation task. It should be worth noting that the model tends to have unreliable representation under the low-budget regime. However, we still lead the performance of all the methods (1.8\% and 0.9\% performance gain over the random baseline with the number of labeled data being 90 in the two datasets). The model tends to have unreliable representation when the labeled dataset is rather small, but we effectively utilize the knowledge of unlabeled instances in the form of feature discrepancy learning to mitigate the prediction bias of the model.

\textbf{Ablation Study.} We additionally conducted an ablation study of our proposed method on CVC-ClinicDB dataset as shown in Table~\ref{table_ablation}. It can be observed that in the early stage of the training, annotating more on the representative samples will lead to better performance but the superiority drops subsequently (w/o uncertainty in the table). The uncertainty sampling benefits the model generalization more in the latter samplings rounds. Our proposed uncertainty-weighted clustering strides a good balance between the uncertain and diverse sampling, which consistently outperforms the sampling methods with only one metric. It is worth noting that the invariant without the feature discrepancy learning is outperformed by our method, indicating that the feature discrepancy learning ensures the robustness of the selection. 

\begin{table}[t]
\centering
\caption{Performance comparison between baselines and our proposed active learning framework on two datasets.}
\label{table1}
\begin{tabular}{c|c|c|c|c|cc} 
\hline
\multicolumn{1}{c|}{\multirow{2}{*}{LN}} &\multirow{2}{*}{Method}& \multicolumn{2}{c|}{CVC-ClinicDB}&\multicolumn{2}{c}{Anon Dataset}\\
\cline{3-6}
\multicolumn{1}{c|}{} & &  mIoU$\uparrow$ & Dice$\uparrow$ & mIoU $\uparrow$ & Dice$\uparrow$    \\
\hline
\multirow{5}{*}{90} &Random	&{0.8388} &0.8225&0.7986&{0.7963}\\
\multirow{3}{*}{} & Entropy	&0.8293&0.8103	&	0.8034&	{0.8023}\\
\multirow{3}{*}{}& ASA &	0.8307	&0.8157	&0.7964	&0.7974\\
\multirow{3}{*}{}& Coreset &0.8252	&0.8099&0.8038	&0.8028	\\
\multirow{3}{*}{}& Ours &\textbf{0.8561}	&\textbf{0.8466}	&\textbf{0.8088}	&\textbf{0.8681}	\\
\hline
\multirow{5}{*}{120} &Random	&{0.8610}	&	0.8517&0.8170&	{0.8182}\\
\multirow{3}{*}{} & Entropy	&0.8622&	0.8562&	0.8180&	{0.8198}\\
\multirow{3}{*}{}& ASA &	0.8438	&	0.8319&0.8133	&0.8151\\
\multirow{3}{*}{}& Coreset &	0.8452	&0.8299	&0.8165&0.8182\\
\multirow{3}{*}{}& Ours &	\textbf{0.8697}	&\textbf{0.8630}&\textbf{0.8213}&\textbf{0.8220}	\\
\hline
\multirow{5}{*}{180} &Random	&{0.8709}	&0.8636	&0.8306&	{0.8355}\\
\multirow{3}{*}{} & Entropy	&0.8810&0.8753	&	0.8295&	{0.8340}\\
\multirow{3}{*}{}& ASA &	0.8703	&	0.8740&0.8178	&0.8185\\
\multirow{3}{*}{}& Coreset &	0.8537	&0.8410	&\textbf{0.8313}	&\textbf{0.8366}\\
\multirow{3}{*}{}& Ours &	\textbf{0.8922}	&\textbf{0.8891}&0.8303	&0.8359	\\
\hline
\end{tabular}
\end{table}

\begin{table}[t]
\centering
\caption{Ablation study results of our proposed method. The performance is evaluated by the mIoU metrics on the CVC-ClinicDB dataset. UNC, CLU, DIS denote uncertainty, clustering, discrepancy learning, respectively.}
\begin{tabular}{ccccccc}
\toprule  
Method & 30 & 60 & 90 & 120 & 150& 180 \\
\midrule  
w/o UNC &75.75&82.63&85.04&85.11&86.73&87.61 \\
w/o CLU &75.75& 79.31&80.46&83.70&85.92&88.48\\
w/o DIS  & 75.75&81.86&85.03&85.93&86.12&86.51\\
Ours &75.75&83.08&85.62&86.97&87.94&89.22\\
\bottomrule 
\end{tabular}
\label{table_ablation}
\end{table}

\section{Conclusion}
In this paper, we proposed a deep active learning framework for medical image
segmentation tasks towards minimizing the annotation cost. The proposed uncertainty-weighted clustering selects uncertain and representative samples to best facilitate the model generalization with a limited annotation budget. The novel feature discrepancy learning also ensures the robustness of the sampling strategy by enhancing the feature representation of the network.

\section{Compliance with Ethical Standards} This research study was conducted retrospectively using human subject data made available in open access by~\cite{cvc-clinicdb}. Ethical approval was not required as confirmed by the license attached with the open-access data.

\section{Acknowledgments} 
This work was supported in part by the National Natural Science Foundation of China (NO.~62322608, No.~82002221, No.~8227242), in part by the Open Project Program of the Key Laboratory of Artificial Intelligence for Perception and Understanding, Liaoning Province (AIPU, No.~20230003), and in part by the Shenzhen Science and Technology Program (NO.~JCYJ20220530141211024), in part by the Sixth Affiliated Hospital of Sun Yat-sen University Start-up Fund for Returnees (No.~R20210217202501975).

{
\bibliographystyle{IEEEbib}
\bibliography{refs}
}

\end{document}